\documentclass[10pt,twocolumn,letterpaper]{article}

\usepackage[pagenumbers]{cvpr} 
\usepackage{multirow}

\definecolor{cvprblue}{rgb}{0.21,0.49,0.74}
\usepackage[pagebackref,breaklinks,colorlinks,allcolors=cvprblue]{hyperref}
\usepackage{algorithm}
\usepackage{algpseudocode}

\def\confName{CVPR}
\def\confYear{2027}

\title{Ground, Guide, Act: Improving VLA Spatial Generalization with \\Multimodal Foundation Models}

\author{Guangting Zheng\thanks{These authors contributed equally.}, Shaoxuan Wang\footnotemark[1], Yu Yuan, Zhipeng Tang, Jiajun Deng, Yanyong Zhang\\
University of Science and Technology of China\\
{\tt\small \{zgt,wangshaoxuan,yyhappier,tangzhipeng\}@mail.ustc.edu.cn, \{dengjj,yanyongz\}@ustc.edu.cn} 
}
\begin{document}
\maketitle
\begin{abstract}
Vision-Language-Action (VLA) models effectively learn action policies from robot demonstrations, but struggle to generalize to changes in object positions and task goals.
General multimodal foundation models (MFMs) offer complementary scene understanding and spatial reasoning, but remain slow and unreliable for direct fine-grained control.
To understand how MFMs can improve VLA spatial generalization, we conduct a systematic capability analysis across multiple VLA architectures using atomic-stage interventions, revealing that many failures stem from poor target grounding and global spatial routing, while local manipulation skills remain largely usable.
Based on these findings, we propose MFM-VLA, a selective collaboration framework in which an MFM provides target grounding and spatial guidance, controllers execute free-space motion, and the VLA retains control over fine-grained manipulation.
We fine-tune the VLA with action-chunk-level dense geometric captions and perturbed-state recovery data to improve local guidance following and recovery from unfamiliar handoff states.
A development-calibrated capability router selects between direct VLA execution and MFM-assisted execution to avoid unnecessary intervention. Across LIBERO-PRO, ALOHA-Sim-PRO, RoboCasa365, and real-world chemistry tasks, MFM-VLA improves success rates by 14.9, 74.0, 3.6, and 30.9 percentage points, respectively, over the corresponding base VLA.
The project page is available at \url{https://tom-zgt.github.io/MFM-VLA}.

\end{abstract}    

\section{Introduction}
\label{sec:introduction}
Vision-Language-Action (VLA) models learn manipulation policies from demonstrations~\cite{rt2,openvla,pi0,pi05,octo}, but struggle with changes in object positions and task goals due in part to limited demonstration coverage~\cite{openx,droid}.
General multimodal foundation models (MFMs) offer complementary capabilities in scene understanding, spatial reasoning, and task planning~\cite{saycan,palme,voxposer}.
Yet direct action generation with these models is slow and unreliable for fine-grained manipulation, while local trajectory corrections alone may not resolve errors in target selection and spatial routing~\cite{hirobot,gpt_vla_assistance}.
This raises a central question: \emph{how can MFMs improve VLA spatial generalization?}

To answer this question, we conduct a capability analysis across multiple VLA architectures, combining counterfactual instruction tests with a five-level Phase-Decoupled Intervention Staircase (PDIS). 
Counterfactual tests provide evidence of visual scene shortcuts: policies often select the original training target despite updated instructions. 
By decomposing manipulation into atomic stages, PDIS further identifies target grounding and global spatial routing as major bottlenecks, while showing that local manipulation skills remain largely usable under appropriate spatial guidance.
For example, guiding the end effector to the correct target increases the target-lifting rate of $\pi_{0.5}$ from 42.3\% to 93.1\%. 
These findings suggest that MFMs should provide target grounding and spatial guidance, while VLAs retain fine-grained manipulation control.

Based on these findings, we propose \textbf{MFM-VLA}, a selective collaboration framework in which an MFM identifies task-relevant manipulation targets and provides spatial guidance, controllers execute free-space motion, and the VLA handles fine-grained manipulation.
To help the VLA use external guidance, we fine-tune it with action-chunk-level dense geometric captions and perturbed-state recovery data, improving local guidance following and recovery from unfamiliar handoff states.
We further introduce a development-calibrated capability router that selects between direct VLA execution and MFM-assisted execution, avoiding unnecessary intervention on tasks the VLA already handles well.

We evaluate MFM-VLA on the LIBERO-PRO benchmark~\cite{libero_pro}, ALOHA-Sim-PRO~\cite{aloha}, RoboCasa365~\cite{robocasa365}, and real-world multi-stage chemistry tasks~\cite{tang2026labdex}, achieving success-rate gains of 14.9, 74.0, 3.6, and 30.9 percentage points over the corresponding base VLAs, respectively.
We also explore an extended system in which the MFM selects between the adapted VLA and task-specific code-based controllers for operations beyond the VLA's skills.

\begin{figure*}[t]
    \centering
    \includegraphics[width=\linewidth]{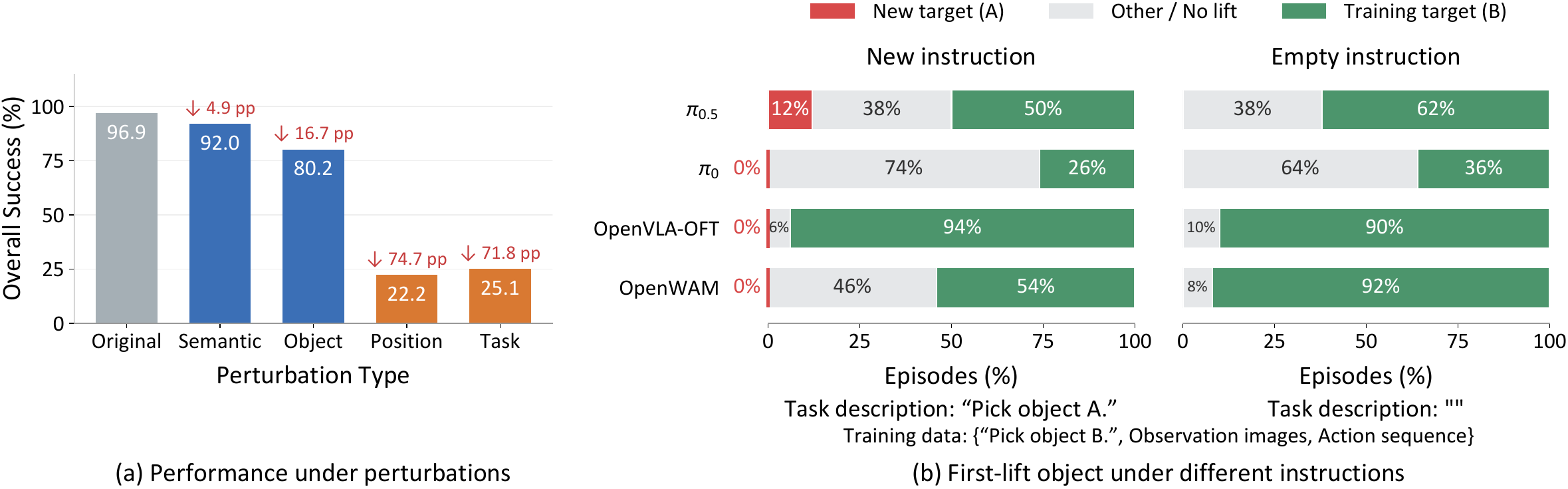}
    \caption{\textbf{Generalization gaps and visual scene shortcuts in pretrained VLAs.}
    (a) $\pi_{0.5}$ degrades sharply under Position and Task perturbations.
    (b) Across four VLA/WAM architectures, updated and empty instructions frequently lead to lifting the original training target, revealing limited instruction sensitivity in familiar scenes.}
    \label{fig:analyse}
\end{figure*}

Our main contributions are summarized as follows:
\begin{itemize}
    \item We systematically analyze multiple pretrained VLAs through counterfactual instruction tests and phase-decoupled interventions, revealing the visual scene shortcuts and identifying the roles of target grounding, spatial routing, and local manipulation in VLA failures.

    \item We propose MFM-VLA, a selective collaboration framework combining dense geometric policy adaptation, perturbed-state recovery, development-calibrated routing, and MFM-based spatial guidance.

    \item We evaluate MFM-VLA on three simulation benchmarks and real-world chemistry tasks, demonstrating its generalization and effectiveness across scenes and tasks.
\end{itemize}

\section{Related Work}
\label{sec:related_work}

\noindent\textbf{Vision-Language-Action Models.}
VLA models have made significant progress in robotic manipulation.
RT-2~\cite{rt2} transfers vision-language knowledge to robot control, while OpenVLA~\cite{openvla} provides an open-source model trained on large-scale robot demonstrations.
$\pi_0$~\cite{pi0} uses flow matching for continuous action generation, and $\pi_{0.5}$~\cite{pi05} extends this approach toward open-world generalization.
OpenVLA-OFT~\cite{openvlaoft} and OpenWAM~\cite{openwam} explore efficient policy fine-tuning and unified world-action modeling, respectively.
Despite these advances, generalization remains challenging when scenes, object positions, or task goals differ from training data~\cite{libero_pro}.
We analyze the roles of target grounding, spatial routing, and local control in these failures to identify where external MFMs can effectively assist VLA policies.

\noindent \textbf{Foundation Models for Robotic Manipulation.}
Language and multimodal foundation models have been used for robotic task planning, executable program generation, and spatial grounding~\cite{saycan,voxposer,code_as_policies,roborefer}.
Other approaches integrate vision-language models (VLMs) into robot policies for spatial guidance and action correction~\cite{omniguide,fpc_vla}, while coding agents generate and refine manipulation skills through execution feedback~\cite{graph_as_policy,cap_x,aspire,enpire,rats}.
PLD further improves pretrained VLAs through reinforcement learning and recovery data generation~\cite{pld}.
Effective collaboration also requires identifying VLA capability gaps, adapting policies to external guidance, and avoiding unnecessary intervention.

Concurrent work explores learned intervention for selective GPT assistance~\cite{call_astra}, visual-trace guidance for VLA policies~\cite{gt_vla}, and hybrid control with pretrained VLAs~\cite{galbot_astra,ebench}.
Other studies investigate direct VLM action generation~\cite{easg} and hierarchical VLM--policy collaboration~\cite{leap_vla}.
Unlike uniform guidance, MFM-VLA combines capability analysis, geometric policy adaptation, and calibrated routing for selective MFM assistance.

\begin{figure*}[]
    \centering

    \begin{subfigure}[t]{0.68\textwidth}
        \centering
        \vspace{0pt}
        \includegraphics[width=\linewidth]{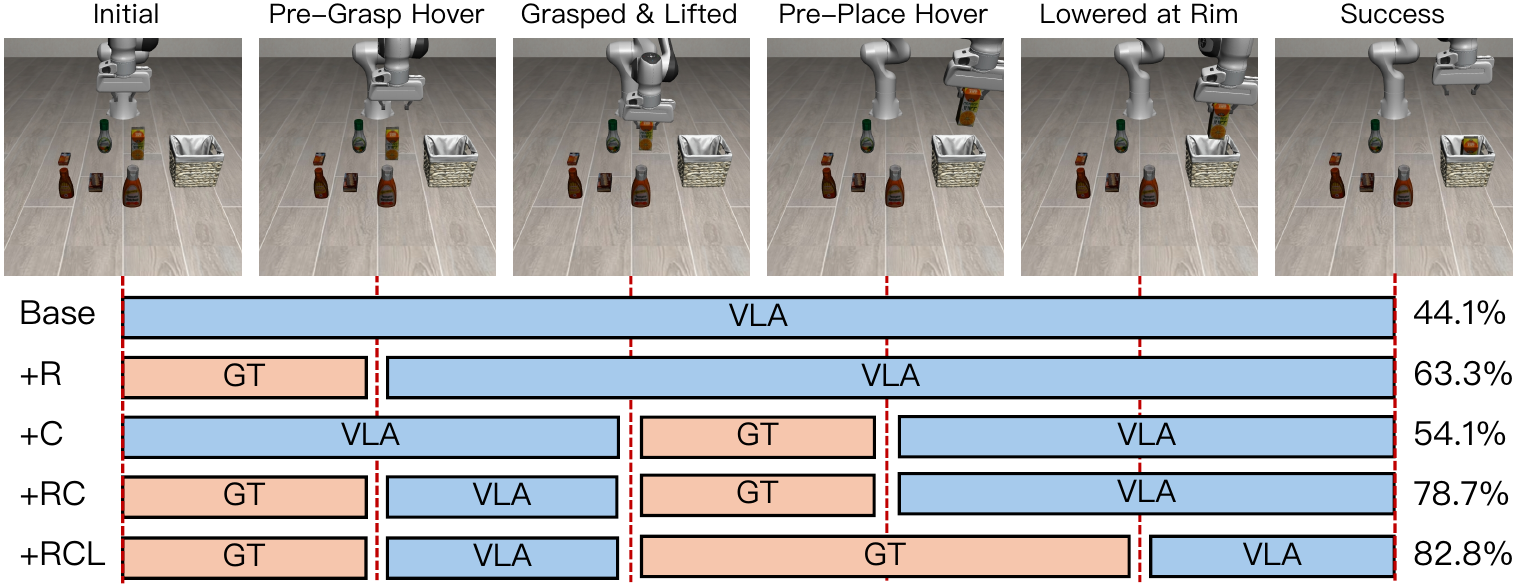}
        \caption{Visualization of atomic stage analysis.}
        \label{fig:pdis_plot}
    \end{subfigure}
    \hfill
    \begin{subfigure}[t]{0.29\textwidth}
        \centering
        \vspace{0pt}
        \setlength{\tabcolsep}{3pt}
        \renewcommand{\arraystretch}{1.15}
        \resizebox{\linewidth}{!}{%
            \begin{tabular}{@{}llccccc@{}}
                \toprule
        Model & Split & \texttt{Base} & \texttt{+R} & \texttt{+C} & \texttt{+RC} & \texttt{+RCL} \\
        \midrule
        $\pi_{0.5}$ & Overall & 44.1 & 63.3 & 54.1 & 78.7 & \textbf{82.8} \\
                    & Pos.    & 13.8 & 42.3 & 43.1 & 68.5 & \textbf{71.5} \\
                    & Task    & 20.0 & 57.7 & 27.7 & 77.7 & \textbf{86.9} \\
        \midrule
        $\pi_0$     & Overall & 30.8 & 32.1 & 35.6 & 53.1 & \textbf{54.4} \\
                    & Pos.    &  3.1 &  5.4 & 16.2 & 28.5 & \textbf{29.2} \\
                    & Task    & 18.5 & 30.8 & 19.2 & 51.5 & \textbf{60.0} \\
        \midrule
        OpenVLA-OFT & Overall & 37.2 & 44.1 & 39.0 & 65.1 & \textbf{66.4} \\
                    & Pos.    & 13.1 & 15.4 & 16.9 & \textbf{53.1} & 50.8 \\
                    & Task    &  7.7 & 39.2 &  7.7 & 57.7 & \textbf{62.3} \\
        \midrule
        OpenWAM     & Overall & 45.4 & \textbf{59.2} & 40.8 & 57.7 & 56.4 \\
                    & Pos.    & 11.5 & 35.4 & 17.7 & \textbf{50.0} & 49.2 \\
                    & Task    & 32.3 & \textbf{55.4} & 20.0 & 50.8 & 46.9 \\
        \bottomrule
            \end{tabular}%
        }
        \caption{Task success across VLA architectures.}
        \label{fig:pdis_table}
    \end{subfigure}

    \caption{
        \textbf{Atomic stage analysis across four VLA architectures}
        ($390$ perturbed episodes per setting).
        Task success is reported in percent.
        Assisting free-space reach (\texttt{+R}) and carry
        (\texttt{+RC}) improves task success without modifying VLA weights.
    }
    \label{fig:pdis_staircase}
\end{figure*}

\section{Capability Analysis of Pretrained VLAs}
\label{sec:analysis}

To understand the limitations of pretrained VLAs and identify where MFM assistance is needed, we investigate their generalization under perturbations, instruction following, and stage-wise manipulation capabilities.

\subsection{Performance Gap Under Perturbations}
\label{sec:analysis_gap}

High success on familiar scenes does not necessarily indicate that a VLA can adapt to new spatial configurations or task goals. To examine this, we evaluate the official $\pi_{0.5}$, trained on LIBERO, on LIBERO-PRO over $8{,}000$ episodes with four types of perturbations: \emph{Semantic} (instruction paraphrases), \emph{Object} (appearance changes), \emph{Position} (object location swaps), and \emph{Task} (changes to the instructed target). 
While Semantic and Object perturbations preserve the required spatial trajectories, Position and Task perturbations require changes in target selection or spatial movement.

As shown in \cref{fig:analyse}(a), $\pi_{0.5}$ maintains high success under Semantic ($92.0\%$) and Object ($80.2\%$) perturbations, but drops sharply under Position ($22.2\%$) and Task ($25.1\%$) perturbations, reaching only $2.4\%$ on LIBERO-10 Position. 
This gap suggests that pretrained VLAs handle surface-level variations better than changes requiring new target selection and spatial trajectories. 
We next investigate whether these failures arise from an over-reliance on familiar visual scenes rather than insufficient local manipulation skills.

\subsection{Visual Scene Shortcuts}
\label{sec:analysis_shortcut}

The performance gap under Task perturbations raises a question: do VLAs fail to manipulate the newly instructed targets, or do they simply ignore updated instructions? 
To investigate this, we conduct counterfactual instruction tests by keeping the scene fixed while changing the instruction. 
Specifically, we compare an updated instruction $\ell_{\mathrm{new}}$, which specifies a new target $o_{\mathrm{new}}$ instead of the original training target $o_{\mathrm{train}}$, with an empty instruction $\ell_{\emptyset}=\texttt{""}$. 
We evaluate $\pi_{0.5}$, $\pi_0$, OpenVLA-OFT, and OpenWAM on LIBERO-Object Task perturbations, with $50$ episodes per condition. 
We measure the lifting rates of both targets and the \emph{Same-Lift Rate}, which measures whether paired rollouts under the two instructions lift the same object first.

As shown in \cref{fig:analyse}(b), OpenVLA-OFT and OpenWAM lift the original training target in $90.0\%$ and $92.0\%$ of empty-instruction episodes, respectively. 
Updated instructions often fail to redirect this behavior: OpenVLA-OFT never lifts the newly instructed target, lifts the original target in $94.0\%$ of episodes, and achieves a $96.0\%$ Same-Lift Rate. 
Similar patterns are observed for $\pi_{0.5}$ on LIBERO-10 ($90.0\%$ agreement) and $\pi_0$ on bimanual ALOHA-Sim ($86.0\%$ agreement).
These results suggest a \textbf{visual scene shortcut}: pretrained VLAs may use familiar visual configurations to retrieve manipulation behaviors, often overriding updated language goals. 
This indicates that improving local action execution alone may be insufficient; explicit target grounding and spatial guidance are also needed.

\begin{figure*}[t]
    \centering
    \includegraphics[width=\textwidth]{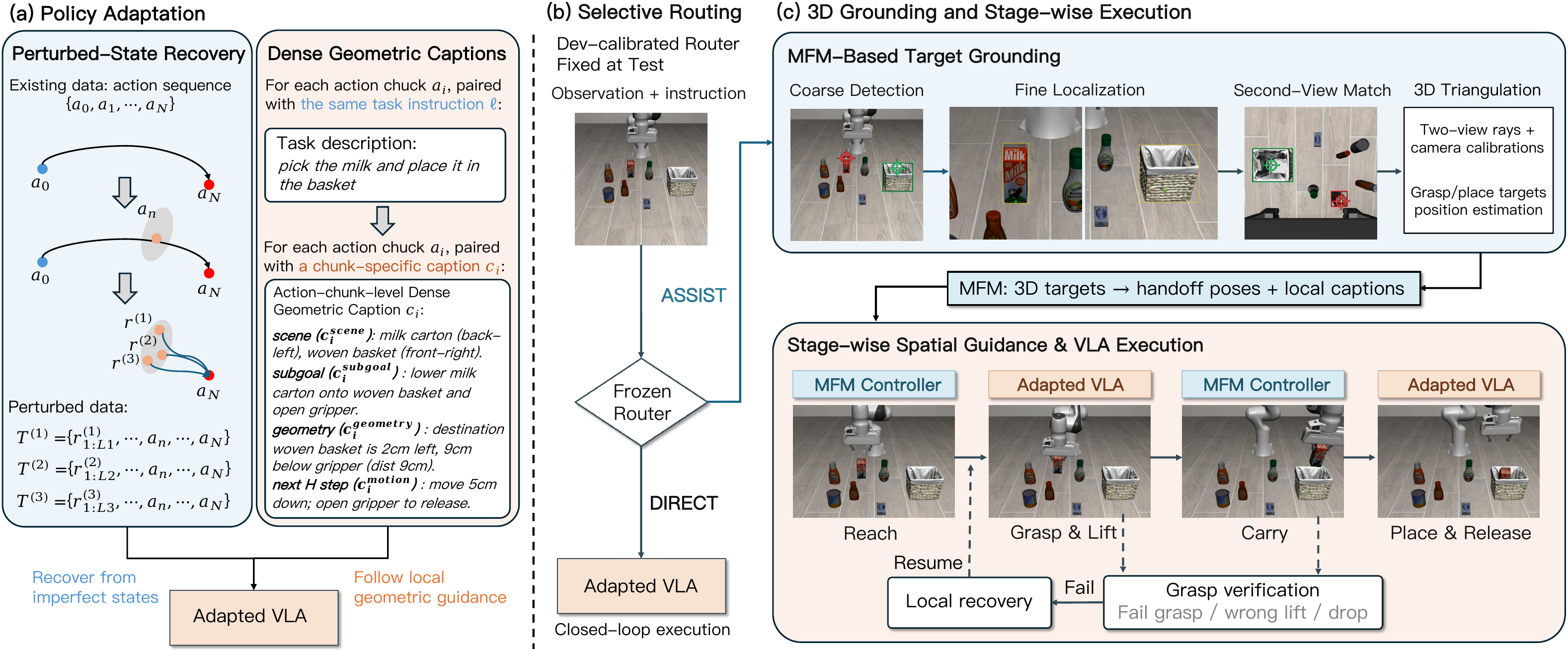}
    \caption{\textbf{Overview of MFM-VLA.}
    Perturbed-state recovery data improve robustness to off-trajectory states, while dense geometric captions enable local guidance following.
    A development-calibrated router selects direct or assisted execution.
    In the illustrated single-arm pipeline, task-aware manipulation-point grounding and two-view triangulation provide spatial targets for controller-executed free-space motion. 
    The adapted VLA performs contact execution, while a separate grasp monitor triggers bounded controller-assisted local recovery.}
    \label{fig:method_overview}
\end{figure*}

\subsection{Atomic Stage Analysis}
\label{sec:analysis_pdis}

The preceding analysis reveals target-selection failures but does not determine whether VLA policies retain local manipulation skills once the correct target is identified. 
To investigate this, we decompose pick-and-place tasks into four atomic stages: \emph{Reach} ($R$), moving the open gripper above the target; \emph{Grasp}, acquiring and lifting the object; \emph{Carry} ($C$), transporting it above the destination; and \emph{Lower/Place} ($L$), lowering and releasing the object. 
We introduce a five-level \emph{Phase-Decoupled Intervention Staircase} (PDIS) to isolate the effects of assistance at different stages: \texttt{Base} uses the VLA throughout; \texttt{+R} assists Reach; \texttt{+C} assists Carry after a successful grasp; \texttt{+RC} assists both Reach and Carry; and \texttt{+RCL} additionally assists Lower/Place.
We use ground-truth spatial guidance with fixed VLA weights to isolate stage-specific assistance benefits without introducing MFM grounding errors. 
All four VLA architectures are evaluated over $390$ perturbed episodes per setting.

As shown in \cref{fig:pdis_staircase}, Reach assistance increases $\pi_{0.5}$'s target-lifting rate from $42.3\%$ to $93.1\%$ under Position perturbations and from $40.0\%$ to $91.5\%$ under Task perturbations. 
Similar improvements across other architectures suggest that local grasping skills remain largely usable once the gripper reaches the correct target. However, successful grasping does not guarantee task completion. Under Position perturbations, $+R$ achieves only $42.3\%$ task success despite a $93.1\%$ target-lifting rate. 
Adding Carry assistance ($+RC$) increases task success to $68.5\%$ on Position and $77.7\%$ on Task, while $+RCL$ further improves these results to $71.5\%$ and $86.9\%$, respectively. 
These results reveal complementary capabilities: MFMs provide target grounding and free-space spatial guidance, while VLAs retain useful local manipulation skills. 
This motivates a collaborative approach that combines MFM spatial reasoning with VLA-based fine-grained control.

\section{Method}
\label{sec:method}
Our analysis identifies target grounding and global spatial routing as major sources of VLA failures, while local manipulation skills remain largely usable. 
Based on this, we propose \textbf{MFM-VLA} (\cref{fig:method_overview}), a selective collaboration framework with three components: (1) \emph{dense geometric policy adaptation} to enable local guidance following and recovery from perturbed states; (2) \emph{development-calibrated capability routing} to determine when to invoke MFMs; and (3) \emph{MFM-based 3D grounding and spatial guidance} to guide free-space motion while retaining VLA control over fine-grained manipulation. We primarily use GPT-6 Astra with $\pi_{0.5}$, and the official $\pi_0$ policy for ALOHA-Sim-PRO.

\subsection{Dense Geometric Policy Adaptation}
\label{sec:method_dense}
Recovery policies have been studied for handling out-of-distribution states in imitation learning~\cite{back_to_manifold}.
To enable the VLA to follow external guidance, we adapt $\pi_{0.5}$ through perturbed-state recovery and action-chunk-level dense geometric captions.

\noindent\textbf{Perturbed-State Recovery.}
Given an original training trajectory $\mathcal{T}=\{a_0,\dots,a_N\}$, we sample an intermediate step $n$ with robot state $s_n$ and perturb the arm to a new pose $s'_n$ without recording. 
A Cartesian controller then drives the arm from $s'_n$ back to $s_n$, recording $L_k$ recovery chunks $\{r_1^{(k)},\dots,r_{L_k}^{(k)}\}$, before replaying the remaining demonstration $\{a_n,\dots,a_N\}$. 
By sampling multiple perturbations ($k=1,2,3$), we construct augmented trajectories: $\mathcal{T}^{(k)}
=
\{r_1^{(k)},\dots,r_{L_k}^{(k)},a_n,\dots,a_N\}$.
Only successful rollouts are retained and added to the original training data, enabling $\pi_\theta$ to recover from perturbed handoff states. 
All recovery data are generated from training demonstrations, without using router-calibration or test episodes.

\noindent\textbf{Dense Geometric Captions.}
Standard VLA fine-tuning pairs all action chunks in an episode with the same task-level instruction $\ell$. 
We instead associate each $H$-step action chunk $a_i$ with a chunk-specific dense geometric caption:
\begin{equation}
c_i=\phi_{\mathrm{dense}}
(o_i,p_i^{\mathrm{eef}},
\hat p_{\mathrm{grasp}},\hat p_{\mathrm{place}}),
\end{equation}
where $c_i$ consists of four fields:
(1) \texttt{scene}, describing the locations of the grasp target and destination;
(2) \texttt{subgoal}, specifying the current manipulation stage;
(3) \texttt{geometry}, encoding the target-relative offset
$\Delta p_i=\hat p_{\mathrm{active}}-p_i^{\mathrm{eef}}$
and its Euclidean distance; and
(4) \texttt{next $H$ steps}, describing short-horizon end-effector displacement and gripper commands.
These fields provide action-specific semantic and geometric conditions for local manipulation.

During training, motion descriptions are generated in two ways: with 50\% probability, they are extracted from the expert's next $H$ actions, including end-effector displacements and gripper commands; otherwise, they are generated using predefined geometric rules based on the current manipulation stage and the 3D displacement from the end effector to the active target.
We fine-tune $\pi_\theta$ on original demonstrations and recovery data, sampling text conditions from $\{\ell,[\ell;c_i],c_i\}$. 
The \textsc{Direct} branch uses the original instruction $\ell$, while the \textsc{Assisted} branch can use geometric captions to follow external spatial guidance.
At inference, motion descriptions are generated from the current manipulation stage and target-relative geometry.

\subsection{Development-Calibrated Capability Routing}
\label{sec:method_router}

As shown in \cref{tab:ablation_dense}, uniform MFM assistance can improve performance on challenging tasks but may interfere with tasks that the adapted VLA already handles well. 
We therefore introduce a development-calibrated router to select between direct and assisted execution.

For each routing unit $\tau$, we evaluate the adapted VLA under both \textsc{Direct} and \textsc{Assisted} execution on a development set $\mathcal{D}_{\mathrm{dev}}$, using $K$ initialization seeds. 
We obtain the corresponding success rates $\hat S_{\mathrm{direct}}(\tau)$ and $\hat S_{\mathrm{assist}}(\tau)$. The routing decision is defined as
\begin{equation}
\small
R(\tau)=
\begin{cases}
\mathrm{ASSISTED},\ 
\hat S_{\mathrm{assist}}(\tau)
-\hat S_{\mathrm{direct}}(\tau)>\delta
\land C(\tau)=1,\\
\mathrm{DIRECT},\ \text{otherwise},
\end{cases}
\end{equation}
where $\tau$ denotes a routing unit, instantiated at either the suite or individual-task level, and $C(\tau)$ indicates whether the task supports the spatial handoff interface. The resulting routing decisions are frozen before evaluation.

In \textsc{Direct} mode, $\pi_\theta$ executes independently using the task instruction $\ell$. 
In \textsc{Assisted} mode, the MFM-based spatial guidance procedure in \cref{sec:method_guidance} is activated.

\subsection{Task-Aware Grounding and Spatial Guidance}
\label{sec:method_guidance}

In \textsc{Assisted} mode, the MFM identifies task-relevant objects and manipulation-specific target points, providing spatial guidance while the adapted VLA retains control over local manipulation.

\noindent\textbf{Two-view 3D Target Grounding.}
To estimate 3D target positions without depth images, we use two calibrated RGB views: a fixed camera and either a second fixed camera or a wrist-mounted camera. Given instruction $\ell$, the MFM identifies task-relevant objects and selects manipulation-specific target points based on object structure and task requirements. For example, opening a bottle requires locating its cap, whereas placing an object into a basket requires grounding an appropriate point inside the basket rather than its object center.
It then refines these points in zoomed crops and establishes semantic correspondences across the two camera views. Camera calibration converts the matched points into viewing rays, whose closest-approach midpoint provides the 3D estimates:
$\hat p_{\mathrm{grasp}}$ and $\hat p_{\mathrm{place}}$.
Geometric consistency checks and fallback rules are detailed in the Appendix.

\noindent\textbf{Stage-wise Spatial Guidance and VLA Execution.}
For single-arm and mobile manipulation, execution follows four stages: \emph{Reach}, \emph{Grasp \& Lift}, \emph{Carry}, and \emph{Place \& Release}. Using the grounded targets, a Cartesian controller moves the open gripper to a pre-grasp handoff pose above the target. 
The adapted VLA $\pi_\theta$ then performs local alignment, descent, grasping, and lifting. After grasp verification, the controller transports the object to a pre-place handoff pose above the destination, where the VLA resumes control for lowering and release. 
The grounded targets also provide the relative geometric conditions for dense captions. 
MFM guidance is queried sparsely, while the controller and VLA execute the corresponding motions in closed loop.

\noindent\textbf{Grasp Verification and Local Recovery.}
A separate monitor checks grasp acquisition and object retention. 
If a failed grasp, wrong-object lift, or drop is detected, the controller opens the gripper, retreats, and returns to a pre-grasp handoff state for another VLA grasp attempt. 
Recovery is bounded by the maximum retry count and episode step budget. 
Platform-specific details are provided in the Appendix.

\section{Experiments}
\label{sec:experiments}

\subsection{Experimental Setup}
\label{sec:exp_setup}

\noindent\textbf{Benchmarks and Evaluation.}
We evaluate MFM-VLA on three simulation benchmarks. 
LIBERO-PRO~\cite{libero_pro} comprises 160 tasks across 16 suites with four perturbation types: Semantic, Object, Position, and Task. 
We use 14,075 trajectories for VLA fine-tuning, including official LIBERO demonstrations, perturbed-state recovery data. 
Initialization seeds 0--4 are used for router calibration, and the resulting routing decisions are frozen before evaluation. 
We report results on the remaining seeds 5--49 ($N=7{,}200$) and the full benchmark ($N=8{,}000$). 
Component and router ablations use seeds 5--9 ($N=800$), a subset of the test set.
RoboCasa365~\citep{robocasa365} evaluates mobile manipulation across 50 tasks: 18 Atomic-Seen, 16 Composite-Seen, and 16 Composite-Unseen. 
We fine-tune on 1,437 trajectories from seen tasks, excluding all Composite-Unseen tasks. 
For each task, we compare Direct and Assist execution on five calibration episodes and select Assisted only when it achieves higher success. 
The resulting task-level router is frozen and applied to all 50 episodes per task, yielding 2,500 evaluation episodes, including the calibration episodes. 
ALOHA-Sim-PRO extends the bimanual Transfer Cube task~\citep{aloha} with Semantic, Object, Position, and Task perturbations, alongside an original setting. 
We evaluate 50 episodes per condition using a consistent 400-step horizon and additionally test target displacements of 0--16\,cm beyond the training region. 
End-to-end task success is the primary metric; object lifting and subgoal completion are reported as other metrics.

\noindent\textbf{Real-world Evaluation.}
We evaluate MFM-VLA on the Franka--XHand platform with a 7-DoF Franka arm and a 12-DoF XHand dexterous hand across 11 glassware manipulation tasks, including grasping, placement, funnel insertion, and multi-stage compositions. 
Scene generalization introduces unseen distractors, while position generalization moves targets to locations absent from the demonstrations. 
We use calibrated fixed and wrist-mounted RGB cameras without depth input and evaluate each task over 20 trials. 
Task and camera details are provided in the Appendix.

\noindent\textbf{Baselines and Implementation.}
MFM-VLA uses GPT-6 Astra with $\pi_{0.5}$ on LIBERO-PRO, RoboCasa365, and real-world tasks, and the official $\pi_0$ checkpoint on ALOHA-Sim-PRO. We compare against the corresponding standalone policies and conduct component ablations to isolate the effects of policy adaptation, routing, and spatial guidance. 
For LIBERO-PRO, the router is calibrated through paired Direct and Assisted evaluations on seeds 0--4, with VLA parameters fixed throughout. 
The resulting routing decisions are frozen before evaluation on seeds 5--49. 
Additional training configurations, control settings, and baseline details are provided in the Appendix.

\subsection{Main Results}
\label{sec:exp_main}

\begin{table}[t]
    \centering
    \caption{\textbf{Main results on LIBERO-PRO.}
    Success rates (\%) on the test set and full benchmark.
    The test set excludes router-calibration episodes, whereas the full benchmark includes all 8,000 episodes. 
    \textit{MFM-VLA + Code Policy} denotes an extended system that adds task-specific code-based controllers for operations beyond the VLA's capabilities. 
    The MFM selects between the adapted VLA policy and these controllers.}
    \label{tab:main_libero}
    \small
    \setlength{\tabcolsep}{2pt}
    \begin{tabular*}{\linewidth}{@{\extracolsep{\fill}}llccccc@{}}
        \toprule
        Evaluation & Method & Sem. & Obj. & Pos. & Task
        & Overall \\
        \midrule
        Test set
        & Base $\pi_{0.5}$
        & 91.7 & 80.2 & 22.1 & 25.1 & 54.8 \\
        ($N=7{,}200$)
        & +\textbf{MFM-VLA}
        & 95.4 & 83.8 & 46.8 & 51.2 & 69.3 \\
        & +Code Policy
        & \textbf{96.7} & \textbf{87.9} & \textbf{79.2} & \textbf{70.9} & \textbf{83.7} \\
        \midrule
        Full benchmark
        & Base $\pi_{0.5}$
        & 92.0 & 80.2 & 22.2 & 25.2 & 54.9 \\
        ($N=8{,}000$)
        & +\textbf{MFM-VLA}
        & 95.4 & 83.8 & 47.5 & 52.6 & 69.8 \\
        & +Code Policy
        & \textbf{96.8} & \textbf{88.1} & \textbf{80.5} & \textbf{72.1} & \textbf{84.4} \\
        \bottomrule
    \end{tabular*}
\end{table}

\noindent\textbf{LIBERO-PRO.}
As shown in \cref{tab:main_libero}, MFM-VLA improves overall success from 54.9\% to 69.8\% on the full benchmark. The largest improvements occur under Position (22.2\% to 47.5\%) and Task perturbations (25.2\% to 52.6\%), while performance under Semantic and Object perturbations is also improved. 
These results demonstrate that MFM assistance particularly benefits tasks requiring changes in target selection and spatial movement.
Some operations, such as sustained plate pushing, remain beyond the capabilities of $\pi_{0.5}$. 
We therefore further explore an extended system, MFM-VLA + Code Policy, in which the MFM selects between the adapted VLA and task-specific code-based controllers. 
This extension improves overall success to 84.4\%, demonstrating the potential of the MFM-VLA framework.

\begin{table}[t]
    \centering
    \caption{\textbf{Zero-shot MFM control on LIBERO-PRO.}
    Success rates (\%) on matched $(\text{suite},\text{task},\text{init})$ keys within each horizon. MFM evaluations use init 0--2, which belong to the router-calibration split. Held-out comparisons are reported in \cref{tab:main_libero}.}
    \label{tab:MFM_direct}
    \small
    \setlength{\tabcolsep}{2pt}
    \begin{tabular*}{\linewidth}{@{\extracolsep{\fill}}lcccccc@{}}
        \toprule
        Method & Horizon & $N$ & Semantic & Position & Task
        & Overall \\
        \midrule
        GPT-6 Astra & $1\times$ & 171
        & 6.9 & 1.8 & 1.8 & 3.5 \\
        Base $\pi_{0.5}$ & $1\times$ & 171
        & 96.6 & 14.3 & 24.6 & 45.6 \\
        \textbf{MFM-VLA} & $1\times$ & 171
        & 93.1 & 55.4 & 73.7 & \textbf{74.3} \\
        \midrule
        GPT-6 Astra & $3\times$ & 179
        & 10.2 & 10.0 & 6.7 & 8.9 \\
        Base $\pi_{0.5}$ & $3\times$ & 179
        & 96.6 & 15.0 & 25.0 & 45.3 \\
        \textbf{MFM-VLA} & $3\times$ & 179
        & 93.2 & 55.0 & 73.3 & \textbf{73.7} \\
        \bottomrule
    \end{tabular*}
\end{table}

\noindent\textbf{Zero-Shot MFM Control.}
\cref{tab:MFM_direct} compares direct GPT-6 Astra control with VLA-based execution on matched episodes. 
GPT-6 Astra achieves only $3.5\%$ success at the native horizon and $8.9\%$ with a $3\times$ horizon, despite target-lifting rates of $32.7\%$ and $54.7\%$, respectively. 
In contrast, MFM-VLA maintains over $73\%$ success, highlighting the importance of retaining VLA-based contact control.

\begin{table}[t]
    \centering
    \caption{\textbf{ALOHA-Sim-PRO bimanual transfer success rates (\%)} under the standard $400$-step horizon ($50$ episodes per condition; \emph{Pert.\ Avg.} averages \emph{Sem.}, \emph{Obj.}, \emph{Pos.}, and \emph{Task}).}
    \label{tab:main_aloha}
    \small
    \setlength{\tabcolsep}{2pt}
    \begin{tabular*}{\linewidth}{@{\extracolsep{\fill}}lcccccc@{}}
        \toprule
        Method & Orig. & Sem. & Obj. & Pos. & Task & Pert.\ Avg. \\
        \midrule
        $\pi_0$ & 70.0 & 40.0 & 54.0 & 0.0 & 0.0 & 23.5 \\
        DOT Policy~\cite{dot_policy} & 88.0 & 88.0 & 16.0 & 2.0 & 0.0 & 26.5 \\
        Diffusion Policy~\cite{diffusion_policy} & 92.0 & 92.0 & 42.0 & 4.0 & 0.0 & 34.5 \\
        ACT~\cite{aloha} & 86.0 & 86.0 & 80.0 & 2.0 & 0.0 & 42.0 \\
        \midrule
        $\pi_0$+\textbf{MFM-VLA} & \textbf{100.0} & \textbf{100.0} & \textbf{98.0} & \textbf{92.0} & \textbf{100.0} & \textbf{97.5} \\
        \bottomrule
    \end{tabular*}
\end{table}

\noindent\textbf{ALOHA-Sim-PRO.}
As shown in \cref{tab:main_aloha}, MFM-VLA improves the average success rate across four perturbation settings from 23.5\% to 97.5\% under the standard 400-step horizon. 
The largest gains occur under Position and Task perturbations, where MFM-VLA achieves 92.0\% and 100.0\% success, respectively. 
These results demonstrate that MFM-assisted spatial guidance also benefits bimanual manipulation under substantial spatial and task changes.

\begin{table}[t]
    \centering
    \caption{\textbf{RoboCasa365 success rates (\%).}
    Comp.-S/U denote seen/unseen composite tasks.
    Code Policy extends MFM-VLA with task-specific controllers for operations beyond the VLA's capabilities.
    Each method is evaluated over 2,500 episodes.}
    \label{tab:main_robocasa}
    \small
    \setlength{\tabcolsep}{5pt}
    \begin{tabular*}{\linewidth}{@{\extracolsep{\fill}}lcccc@{}}
        \toprule
        Method & Atomic & Comp.-S & Comp.-U & Overall \\
        \midrule
        Diffusion Policy & 15.7 & 0.2 & 1.3 & 6.1 \\
        $\pi_0$ & 34.6 & 6.1 & 1.1 & 14.8 \\
        $\pi_{0.5}$ & 39.6 & 7.1 & 1.2 & 16.9 \\
        \midrule
        $\pi_{0.5}$ + \textbf{MFM-VLA} & 39.3 & 17.1 & 2.4 & 20.5 \\
        +Code Policy & \textbf{51.8} & \textbf{18.1} & \textbf{18.3} & \textbf{30.3} \\
        \bottomrule
    \end{tabular*}
\end{table}

\begin{figure*}[t]
    \centering
    \includegraphics[width=\textwidth]{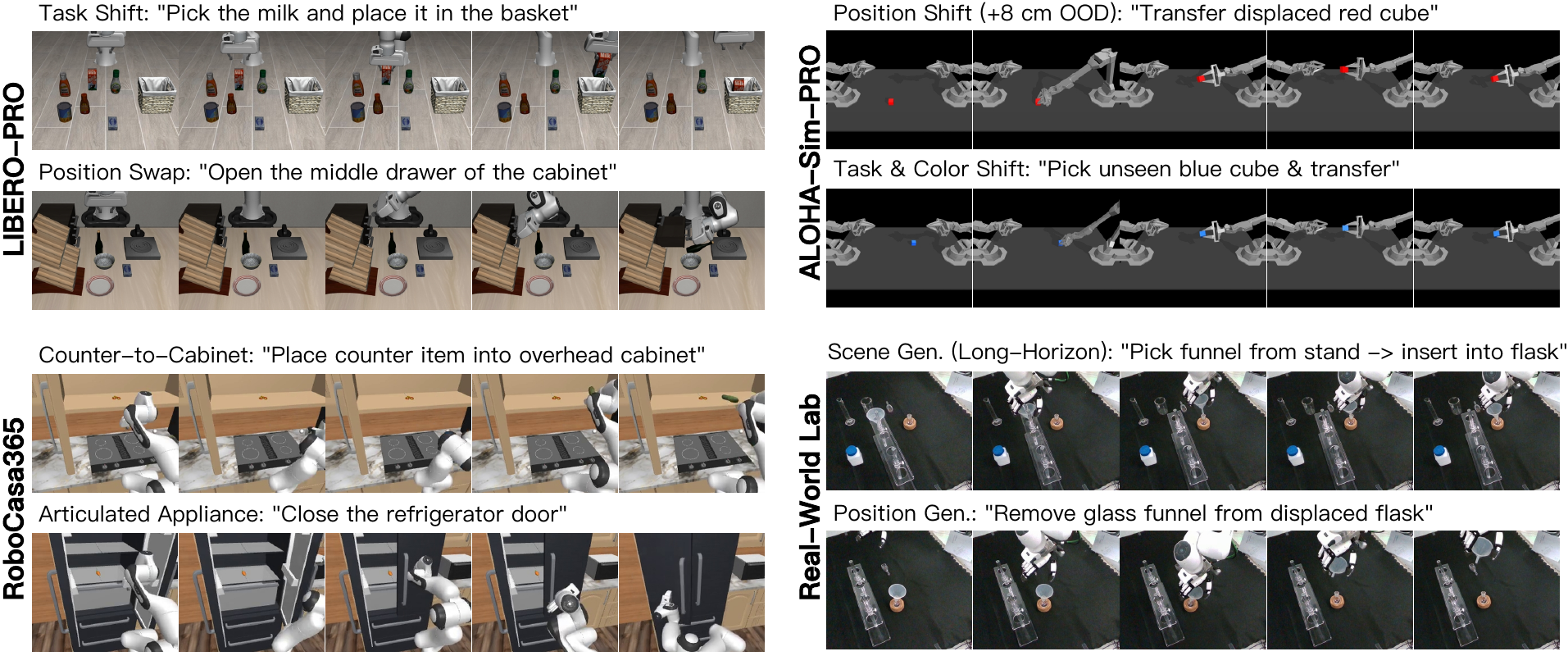}
    \caption{\textbf{Qualitative results across simulation and real-world tasks.}
    We show eight successful executions on LIBERO-PRO, RoboCasa365, ALOHA-Sim-PRO, and the real-world laboratory platform.
    Each row presents successive execution frames from left to right, covering task and position changes, articulated-object manipulation, bimanual transfer, and transparent glassware manipulation.}
    \label{fig:qualitative}
\end{figure*}
\begin{table*}[t]
    \centering
    \scriptsize
    \setlength{\tabcolsep}{3pt}
    \caption{\textbf{Real-world success rates (\%).}
    Scene introduces unseen distractors; Position changes target locations. 
    Each task is evaluated over 20 trials.
    Avg.\ is the unweighted mean across all tasks and both settings within each task group.}
    \label{tab:main_realworld}
    \resizebox{\textwidth}{!}{%
    \begin{tabular}{l*{24}{c}}
        \toprule
        & \multicolumn{17}{c}{\textbf{Atomic tasks}}
        & \multicolumn{7}{c}{\textbf{Multi-stage tasks}} \\
        \cmidrule(lr){2-18}\cmidrule(lr){19-25}

        \multirow{2}{*}{\textbf{Method}}
        & \multicolumn{2}{c}{Pick Beaker}
        & \multicolumn{2}{c}{Pick Cylinder}
        & \multicolumn{2}{c}{Balance}
        & \multicolumn{2}{c}{Place}
        & \multicolumn{2}{c}{Pick Funnel}
        & \multicolumn{2}{c}{Remove Funnel}
        & \multicolumn{2}{c}{Place Funnel}
        & \multicolumn{2}{c}{Return Funnel}
        & \multirow{2}{*}{\textbf{Avg.}}
        & \multicolumn{2}{c}{Mid Balance}
        & \multicolumn{2}{c}{Place Funnel}
        & \multicolumn{2}{c}{Return Funnel}
        & \multirow{2}{*}{\textbf{Avg.}} \\
        \cmidrule(lr){2-3}\cmidrule(lr){4-5}
        \cmidrule(lr){6-7}\cmidrule(lr){8-9}
        \cmidrule(lr){10-11}\cmidrule(lr){12-13}
        \cmidrule(lr){14-15}\cmidrule(lr){16-17}
        \cmidrule(lr){19-20}\cmidrule(lr){21-22}
        \cmidrule(lr){23-24}

        & Scene & Position
        & Scene & Position
        & Scene & Position
        & Scene & Position
        & Scene & Position
        & Scene & Position
        & Scene & Position
        & Scene & Position
        &
        & Scene & Position
        & Scene & Position
        & Scene & Position
        & \\
        \midrule

        $\pi_{0.5}$
        & 0.0  & 0.0
        & 0.0  & 0.0
        & 85.0 & 25.0
        & 0.0  & 45.0
        & 15.0 & 0.0
        & 30.0 & 0.0
        & 15.0 & 0.0
        & 30.0 & 0.0
        & 15.3
        & 0.0  & 0.0
        & 5.0  & 0.0
        & 0.0  & 0.0
        & 0.8 \\

        +\textbf{MFM-VLA}
        & 80.0 & 90.0
        & 0.0  & 30.0
        & 95.0 & 40.0
        & 10.0 & 50.0
        & 80.0 & 75.0
        & 95.0 & 85.0
        & 40.0 & 30.0
        & 25.0 & 45.0
        & \textbf{54.4}
        & 80.0  & 55.0
        & 10.0 & 15.0
        & 15.0 & 15.0
        & \textbf{31.7} \\
        \bottomrule
    \end{tabular}%
    }
\end{table*}
\noindent\textbf{RoboCasa365.}
As shown in \cref{tab:main_robocasa}, MFM-VLA improves overall success from 16.9\% to 20.5\%, with a larger gain on Composite-Seen tasks (7.1\% to 17.1\%). 
Performance on Atomic tasks remains nearly unchanged (39.6\% to 39.3\%), while Composite-Unseen success increases only from 1.2\% to 2.4\%. 
These results suggest that spatial guidance helps combine learned manipulation skills in familiar tasks, but cannot fully compensate for missing low-level skills in unseen task compositions. 
Adding task-specific code-based controllers further improves Composite-Unseen success to 18.3\% and overall success to 30.3\%.

\noindent\textbf{Real-world Manipulation.}
As shown in \cref{tab:main_realworld}, MFM-VLA improves the average success rate from 15.3\% to 54.4\% on atomic tasks and from 0.8\% to 31.7\% on multi-stage tasks, averaged across Scene and Position settings. 
Notable gains include funnel grasping and removal under scene perturbations, as well as beaker grasping and multi-stage manipulation under position changes. 
These results demonstrate the effectiveness of MFM-assisted spatial guidance in real-world manipulation involving transparent objects and multi-stage operations.

\noindent\textbf{Qualitative Results.}
\cref{fig:qualitative} presents representative executions across LIBERO-PRO, RoboCasa365, ALOHA-Sim-PRO, and real-world laboratory tasks. 
The examples cover target and position changes, articulated-object manipulation, bimanual transfer, and transparent glassware operations, illustrating MFM-VLA's applicability across different embodiments and manipulation settings.

\subsection{Ablation Studies}
\label{sec:exp_ablation}

Unless otherwise specified, ablations use the 800-episode LIBERO-PRO subset (init 5--9), excluding router-calibration episodes. 
Under the default suite-level router, 550 episodes use Direct mode, and 250 use Assisted mode.

\begin{table}[t]
    \centering
\caption{\textbf{MFM-VLA component ablations.}
Success rates (\%) on 800 LIBERO-PRO episodes.
\textit{w/o DGC} uses perturbed-state recovery data
but no dense geometric captions.}
    \label{tab:ablation_dense}
    \small
    \setlength{\tabcolsep}{3pt}
    \resizebox{\linewidth}{!}{%
        \begin{tabular}{lcccccc}
            \toprule
            Model & Router & Sem. & Obj. & Pos. & Task & Overall \\
            \midrule
            $\pi_{0.5}$ & No & 92.5 & 81.0 & 20.5 & 28.0 & 55.5 \\
            $\pi_{0.5}$ & Always & 66.0 & 50.0 & 37.5 & 58.0 & 52.9 \\
            Adapted $\pi_{0.5}$ & No & \textbf{95.5} & \textbf{85.0} & 39.5 & 30.5 & 62.6 \\
            \midrule
            $\pi_{0.5}$ & Yes & 92.5 & 81.0 & 41.5 & 57.0 & 68.0 \\
            Adapted $\pi_{0.5}$ (w/o DGC) & Yes & \textbf{95.5} & 81.0 & 54.5 & 56.5 & 71.9 \\
            Adapted $\pi_{0.5}$ & Yes & \textbf{95.5} & \textbf{85.0} & \textbf{56.5} & \textbf{63.5} & \textbf{75.1} \\
            \bottomrule
        \end{tabular}%
    }
\end{table}

\noindent\textbf{Component Ablation.}
\cref{tab:ablation_dense} evaluates the contributions of policy adaptation and selective routing. 
Uniform MFM assistance improves Position and Task performance but degrades Semantic and Object performance, reducing success from 55.5\% to 52.9\%. 
This motivates selective rather than uniform intervention. Dense geometric adaptation alone improves overall success to 62.6\%, while adding the router to the base policy raises it to 68.0\%. 
Combining routing with perturbed-state recovery further improves success to 71.9\%, and adding dense geometric captions increases it to 75.1\%. 
These results demonstrate the complementary benefits of policy adaptation and selective assistance.

\begin{table}[t]
    \centering
    \caption{\textbf{Comparison of routing strategies on LIBERO-PRO.}
    Success rates and assistance rates on 800 episodes
    (init 5--9).}
    \label{tab:ablation_router}
    \small
    \setlength{\tabcolsep}{5pt}
    \begin{tabular*}{\linewidth}{@{\extracolsep{\fill}}lcc@{}}
        \toprule
        Routing Strategy & Success rate (\%) & Assistance rate (\%)\\
        \midrule
        All Direct & 62.6 & 0.0 \\
        All Guided & 62.3 & 100.0 \\
        Perturbation-level & 74.1 & 50.0 \\
        Suite-level & 75.1 & 31.2 \\
        Task-level & \textbf{78.5} & \textbf{27.5} \\
        \bottomrule
    \end{tabular*}
\end{table}

\begin{table}[t]
    \centering
    \caption{\textbf{Spatial guidance ablations.}
    Success rates (\%) on the 250 Assisted episodes
    and all 800 evaluation episodes.}
    \label{tab:ablation_guidance}
    \small
    \setlength{\tabcolsep}{5pt}
    \begin{tabular*}{\linewidth}{@{\extracolsep{\fill}}lcc@{}}
        \toprule
        Variant & Assisted & Overall \\
        \midrule
        Adapted Direct & 33.6 & 62.6 \\
        Single-view grounding & 45.6 & 66.4 \\
        Without carry guidance & 62.0 & 71.5 \\
        Without grasp retries & 69.2 & 73.8 \\
        MFM-VLA & \textbf{73.6} & \textbf{75.1} \\
        \bottomrule
    \end{tabular*}
\end{table}

\noindent\textbf{Router Strategy Ablation.}
\cref{tab:ablation_router} compares routing strategies at different granularities.
All-Direct and All-Guided achieve 62.6\% and 62.3\%, respectively, while perturbation-level routing reaches 74.1\%.
Our suite-level router achieves 75.1\% success with only 31.2\% MFM calls.
Although task-level routing further improves success to 78.5\% and reduces MFM calls to 27.5\%, it requires more calibration data to make reliable task-specific decisions.
With limited per-task calibration data, its routing decisions exhibit higher variance and are more sensitive to individual calibration outcomes.
In contrast, suite-level routing aggregates feedback across tasks, enabling more stable decisions with fewer samples.
We thus adopt suite-level routing to balance
performance and stability.

\noindent\textbf{Spatial Guidance Ablation.}
\cref{tab:ablation_guidance} evaluates guidance components, reporting success on the 250 Assisted episodes and overall success across all 800 episodes. 
Replacing two-view triangulation with single-view grounding reduces Assisted success from 73.6\% to 45.6\%, highlighting the importance of accurate 3D localization. 
Removing Carry guidance further reduces success to 62.0\%, while disabling grasp retries reduces it to 69.2\%. 
These results confirm the contributions of accurate grounding, free-space guidance, and local recovery to successful manipulation.

\begin{table}[t]
    \centering
    \caption{\textbf{MFM backbone ablation on LIBERO-PRO.}
    We report median horizontal $(x,y)$ grounding error and
    the percentage of estimates within 6\,cm of the ground-truth
    target position, computed over samples with available estimates.
    Success rates (\%) are reported on the 250 Assisted episodes
    and all 800 episodes.}
    \label{tab:ablation_MFM}
    \small
    \setlength{\tabcolsep}{2pt}
    \begin{tabular*}{\linewidth}{@{\extracolsep{\fill}}lcccc@{}}
        \toprule
        MFM backbone
        & $e_{xy}$ (cm)$\downarrow$
        & $\leq 6$\,cm (\%)$\uparrow$
        & Assisted
        & Overall \\
        \midrule
        Kimi K3
        & 22.94 & 8.7 & 29.6 & 61.4 \\
        Qwen3-VL-30B
        & 2.25 & 63.0 & 38.0 & 64.0 \\
        Gemini 4 Argon
        & 2.75 & 76.1 & 58.4 & 70.4 \\
        Grok 4.7
        & 0.92 & 91.3 & 59.2 & 70.6 \\
        GPT-5.6
        & 0.57 & 93.5 & 60.4 & 71.0 \\
        GPT-6 Astra
        & \textbf{0.16} & \textbf{97.3}
        & \textbf{73.6} & \textbf{75.1} \\
        \bottomrule
    \end{tabular*}
\end{table}

\noindent\textbf{MFM Backbones.}
\cref{tab:ablation_MFM} compares six MFM backbones with
the adapted VLA and execution pipeline fixed.
GPT-6 Astra achieves the lowest median horizontal grounding error (0.16\,cm), the highest $\leq 6$\,cm hit rate (97.3\%),
and the highest Assisted success rate (73.6\%).
These results show that MFM-VLA accommodates different MFM backbones and suggest that more accurate task-relevant grounding contributes to better manipulation performance.

\begin{table}[t]
    \centering
    \caption{\textbf{Target-selection diagnostic after adaptation and guidance.}
    Results (\%) on an independent 200-episode diagnostic set across four Task-perturbation suites.
    $o_{\mathrm{new}}$ denotes the instructed target and $o_{\mathrm{train}}$ the original training target.
    }
    \label{tab:shortcut_reaudit}
    \small
    \setlength{\tabcolsep}{3pt}
    \begin{tabular*}{\linewidth}{@{\extracolsep{\fill}}lccc@{}}
        \toprule
        Method & Success $\uparrow$
        & Lift $o_{\mathrm{new}}$ $\uparrow$
        & Lift $o_{\mathrm{train}}$ $\downarrow$ \\
        \midrule
        Base $\pi_{0.5}$ & 22.5 & 33.0 & 40.0 \\
        Adapted Direct & 27.5 & 35.0 & 35.5 \\
        Base + Guidance & 61.0 & 75.5 & 8.5 \\
        MFM-VLA & \textbf{73.5} & \textbf{81.5} & \textbf{2.5} \\
        \bottomrule
    \end{tabular*}
\end{table}

\noindent\textbf{Shortcut Mitigation.}
To examine whether MFM-VLA mitigates the visual scene shortcuts identified in \cref{sec:analysis_shortcut}, we repeat the counterfactual instruction test on four Task-perturbation suites. 
As shown in \cref{tab:shortcut_reaudit}, dense adaptation alone provides limited improvement in instructed-target lifting (33.0\% to 35.0\%). 
Spatial guidance increases it to 75.5\% and reduces incorrect lifting of the original training target from 40.0\% to 8.5\%. 
Combining guidance with policy adaptation further improves instructed-target lifting to 81.5\% and reduces original-target lifting to 2.5\%. 
These results show that spatial guidance helps redirect target selection, while policy adaptation improves its execution.

\section{Conclusion}

We investigate how MFMs can improve the spatial generalization of pretrained VLAs and make two key observations. First, systematic analysis reveals that pretrained VLAs often rely on visual scene shortcuts, leading to incorrect target selection and spatial routing, while local manipulation skills remain largely usable. Second, indiscriminate MFM assistance can interfere with tasks that VLAs already handle well. Based on these findings, we propose MFM-VLA, a selective collaboration framework that combines dense geometric policy adaptation, development-calibrated routing, and MFM-based spatial guidance. Experiments across three simulation benchmarks and real-world chemistry tasks demonstrate improved spatial generalization and reduced shortcut-driven failures while preserving the manipulation capabilities of pretrained VLAs. 
We hope this work provides a practical direction for combining MFM spatial reasoning with learned VLA manipulation skills.

\section*{LLM Usage Statement}
We used Cursor to assist with code editing and debugging, and ChatGPT for language polishing, grammar checking, and improving manuscript clarity.
All AI-assisted code and text were reviewed and revised by the authors.
{
    \small
    \bibliographystyle{ieeenat_fullname}
    \bibliography{main}
}

\clearpage
\setcounter{page}{1}
\maketitlesupplementary

\section{Implementation Details}
\label{sec:supp_impl}

\subsection{Inference Procedure}
\label{sec:supp_inference}

\newcounter{suppalgorithm}
\renewcommand{\thesuppalgorithm}{S\arabic{suppalgorithm}}
\crefname{suppalgorithm}{Algorithm}{Algorithms}
\Crefname{suppalgorithm}{Algorithm}{Algorithms}
\crefformat{suppalgorithm}{Algorithm~#2#1#3}
\Crefformat{suppalgorithm}{Algorithm~#2#1#3}

\noindent\begin{minipage}{\linewidth}
        \refstepcounter{suppalgorithm}\label{alg:supp_inference}
        \small\noindent\textbf{Algorithm \thesuppalgorithm: MFM-VLA Inference}\par
        \vspace{2pt}\hrule\vspace{4pt}
        \footnotesize
        \noindent\textbf{Input:} Task instruction $\ell$, routing unit $\tau$, frozen router $R$, adapted VLA $\pi_\theta$, calibrated RGB cameras.\par
        \vspace{3pt}
        \begin{tabular}{@{}r@{\hspace{0.5em}}p{0.88\linewidth}@{}}
            1 & \textbf{if} $R(\tau)=\mathrm{DIRECT}$ \textbf{then} execute $\pi_\theta$ with $\ell$; \textbf{return}. \\
            2 & Acquire fixed-view image $I_a$. \\
            3 & Ground the grasp target and destination in $I_a$ with coarse-to-fine MFM queries. \\
            4 & Plan and execute a free-space wrist survey move; acquire second view $I_b$. \\
            5 & \textbf{for} each target $x\in\{\mathrm{grasp},\mathrm{place}\}$ \textbf{do} \\
            6 & \quad Match the task-relevant contact point across $I_a$ and $I_b$. \\
            7 & \quad Triangulate, check visibility and ray gap, and reobserve if necessary. \\
            8 & \quad Set $\hat{\mathbf p}_x$ to the accepted 3D point, or $\varnothing$ if invalid. \\
            9 & \textbf{end for} \\
            10 & \textbf{if} $\hat{\mathbf p}_{\rm grasp}\ne\varnothing$ \textbf{then} estimate support height and execute Cartesian pre-grasp handoff. \\
            11 & \textbf{if} the grasp target is valid \textbf{then} enable grounded dense captions. \\
            12 & Execute VLA-based local alignment, grasp, and lift. \\
            13 & \textbf{if} a grasp is detected and the grasp reference is available \textbf{then} visually verify the held object; retry a wrong grasp at most twice. \\
            14 & \textbf{if} grasp verified and place target valid \textbf{then} execute the platform-specific carry policy:
    LIBERO-PRO: allow a brief VLA carry attempt, then
    switch to Cartesian transport when triggered;
    RoboCasa365: use Cartesian transport immediately. \\
            15 & \textbf{if} a carry drop occurs \textbf{then} re-ground and retry at most once. \\
            16 & Continue VLA execution for local placement and release, subject to the episode step budget. \\
        \end{tabular}
        \vspace{3pt}\hrule
\end{minipage}
\par\medskip

\cref{alg:supp_inference} summarizes the inference procedure of MFM-VLA. 
A frozen capability router first determines whether an episode requires external assistance. 
Direct episodes are executed entirely by the adapted VLA. 
For Assisted episodes, the MFM localizes task-relevant targets from two sequential RGB observations, after which geometric checks determine whether Cartesian handoffs are reliable. 
The controller handles free-space movement where valid targets are available, while the VLA retains control over contact-rich manipulation. 
Visual verification and bounded recovery are used to handle incorrect grasps and dropped objects.

\paragraph{Capability routing.}
The router compares Direct and Assisted success rates measured on a development set and selects Assisted execution only when the estimated gain exceeds $\delta=0.05$ (5 percentage points) and the task supports the handoff interface ($C(\tau)=1$).
We use suite-level routing with $K=5$ calibration episodes per task by default and freeze the resulting routing decisions before evaluation.
A Direct decision executes $\pi_\theta$ with the original instruction $\ell$ without invoking the MFM.

\paragraph{Target grounding and view acquisition.}
For Assisted execution, the MFM first identifies the object to grasp and its destination in the fixed RGB view. 
A full-frame query proposes candidates, and a subsequent zoomed-crop query resolves the intended object and its base-contact pixel. 
These first-view observations define calibrated rays for planning a free-space survey move that places the wrist camera where the candidate targets are likely to be visible. 
Only after this move do we acquire the wrist RGB image. 
Conditioned on the selected objects in the first view, the MFM then identifies the corresponding manipulation points in the two views, rather than independently selecting a new object in the wrist view.

\paragraph{Two-view reconstruction.}
A pixel $\mathbf u_c$ in camera $c$ defines the viewing ray
\begin{equation}
\mathbf r_c(s)=\mathbf o_c+s\mathbf d_c,\qquad
\mathbf d_c=\frac{R_cK_c^{-1}[\mathbf u_c;1]}{\|R_cK_c^{-1}[\mathbf u_c;1]\|_2},
\end{equation}
where $K_c$ denotes camera intrinsics and $(\mathbf o_c,R_c)$ its camera-to-world pose. 
For matched pixels, we find the closest points on the two viewing rays by solving
\[
(s^\star,t^\star)
=
\arg\min_{s,t\geq 0}
\|r_a(s)-r_b(t)\|_2^2.
\]
We take their midpoint as the 3D estimate and their distance as the geometric consistency gap:
\begin{equation}
\hat{p} =
\frac{r_a(s^\star)+r_b(t^\star)}{2},
\quad
g = \|r_a(s^\star)-r_b(t^\star)\|_2.
\end{equation}
If the target is clipped, the wrist camera is recentered; if it is not visible, the hand is raised by $15\,\mathrm{cm}$ once. 
A gap above $1\,\mathrm{cm}$ triggers up to two additional recentering attempts, retaining the valid view with the smallest gap. 
Estimates are accepted for $g\le 2.5\,\mathrm{cm}$, or $3.0\,\mathrm{cm}$ for articulated handles.

\paragraph{Support height and fallback.}
The support height is estimated without depth using the fixed-view ray through the object's base-contact pixel. 
Given the triangulated horizontal contact-point location $\hat{p}_{xy}$ and the base-contact ray $(o_a,d_{\mathrm{base}})$, we find the horizontally closest point on the ray by constraining the projection parameter to be nonnegative:
\begin{equation}
\begin{aligned}
s_\perp &=
\max\left\{
0,\,
\frac{
d_{\mathrm{base},xy}^{\top}
(\hat{p}_{xy}-o_{a,xy})
}{
\|d_{\mathrm{base},xy}\|_2^2
}
\right\},\\
z_{\mathrm{sup}} &=
o_{a,z}+s_\perp d_{\mathrm{base},z}.
\end{aligned}
\end{equation}
If a target has no accepted two-view estimate, its precision-dependent Cartesian handoff is skipped and the adapted VLA continues execution.

\paragraph{Geometric captions and stage-wise control.}
When grounding is available, the adapted VLA receives dense captions describing the scene, current subgoal, gripper-to-target geometry, and short-horizon motion. 
The Cartesian controller moves the open gripper to a pre-grasp handoff pose, after which the VLA performs local alignment, descent, grasping, and lifting. 
Following a verified grasp, carry is primarily handled by the Cartesian controller, which transports the grasped object to a pre-place handoff pose above the destination.
On LIBERO-PRO Assisted episodes (\texttt{eval\_g5.py}, mode \texttt{g5j}), after a confirmed grasp the VLA carries first while a monitor tracks the horizontal end-effector--to--place distance
$d_t=\lVert p_t^{xy}-\hat{\mathbf{p}}_{\mathrm{place}}^{xy}\rVert_2$.
Controller takeover is triggered by any of the following conditions from the frozen \texttt{g5j} place config:
\begin{itemize}
    \item \textbf{Insufficient progress (brief window).}
    After exactly $W_0{=}20$ action steps, if $d_0-d_{W_0}<3\,\mathrm{cm}$.
    \item \textbf{Stall.}
    After the brief window, if over any subsequent window of $W{=}10$ steps the reduction $d_{t-W}-d_t$ is less than $1\,\mathrm{cm}$ while $d_t>10\,\mathrm{cm}$ (near-target horizontal stall is ignored so that lowering/placing is not interrupted).
    \item \textbf{Timeout.}
    After $T_{\mathrm{carry}}{=}40$ action steps without arriving within $4\,\mathrm{cm}$ of $\hat{\mathbf{p}}_{\mathrm{place}}$.
    \item \textbf{Premature gripper opening.}
    If the VLA issues an open-gripper command while $d_t>8\,\mathrm{cm}$.
\end{itemize}
Otherwise, if the VLA opens near the target or reaches $d_t\le 4\,\mathrm{cm}$, control remains with the VLA for Place \& Release (with a free-space release fallback if it has not opened within $30$ steps after hand-back).
On RoboCasa365, controller transport begins immediately once the lift and destination-validity conditions are met.
The VLA then performs local placement and release. Handoff offsets are taken from training demonstrations (e.g., the LIBERO pre-grasp offset is $[+2.6,+0.4,+9.0]\,\mathrm{cm}$ from the object top).

\paragraph{Verification and recovery.}
After a detected grasp-and-lift event, the MFM compares the held object against the reference crop cached during initial grounding, using current fixed and wrist RGB
observations. 
If the wrong object is held, the controller opens the gripper, rises by 5 cm, and returns to a closer pre-grasp handoff pose for another VLA grasp attempt.
The cached reference crop is reused for subsequent grasp verification, with at most $M_{\mathrm{retry}}=2$ re-grasps.
A drop during transport instead triggers re-grounding and at most one carry recovery. 
All actions remain subject to the episode step budget.

\subsection{Benchmark Setup and Inference}
\label{sec:supp_benchmarks}

\paragraph{LIBERO-PRO.}
LIBERO-PRO~\cite{libero_pro} contains 160 tasks across 16 suites, covering four perturbation types: Semantic (instruction paraphrases), Object (appearance changes), Position (object displacement and swaps), and Task (changes in instructed targets).
We fine-tune $\pi_{0.5}$ on 14,075 trajectories, including 6,350 official demonstrations, 5,597 perturbed-state recovery rollouts, and 2,128 spatial augmentations, for 10,000 steps.
We evaluate 50 initializations per task ($N=8{,}000$), using \texttt{init 0--4} for router calibration and \texttt{init 5--49} for held-out evaluation.

We use GPT-6 Astra~\cite{astra} as the external MFM and a frozen suite-level router with $\delta=0.05$, calibrated on five initializations per task.
The router selects Direct execution for 11 suites and Assisted execution for the remaining five.
The adapted VLA receives $224\times224$ RGB observations and replans every five action steps.
Episode horizons follow the benchmark settings: 220 steps for Spatial, 280 for Object, 300 for Goal, and 520 for LIBERO-10 tasks.
In Assisted mode, the MFM grounds task-relevant targets from RGB observations, while the Cartesian controller and VLA execute the corresponding spatial handoffs and local manipulation.
\begin{table*}[t]
    \centering
    \small
    \setlength{\tabcolsep}{5pt}
    \caption{\textbf{Grounding, verification, and retry ablations}
    on LIBERO-PRO (\texttt{init 5--9};
    Assisted $N{=}250$, Overall $N{=}800$).
    The target-ID oracle is a simulation-only diagnostic for grasp verification. Success rates are percentages. 
    For GPT-6 Astra, the two retry settings share the same visual verifier and differ in the maximum number
of re-grasp attempts.}
    \label{tab:supp_gdino_2x2}
    \begin{tabular*}{\textwidth}
        {@{\extracolsep{\fill}}llcc@{}}
        \toprule
        3D Grounding & Verification / Retry Setting
        & Assist. & Overall \\
        \midrule
        GroundingDINO (top-edge)
        & No MFM verify (proprio.\ only)
        & 10.0 & 55.3 \\
        GroundingDINO (box center)
        & No MFM verify (proprio.\ only)
        & 16.4 & 57.3 \\
        GroundingDINO (box center)
        & Live MFM visual (\texttt{Qwen3-VL-30B})
        & 17.2 & 57.5 \\
        GroundingDINO (box center)
        & Sim target-ID oracle
        & 17.2 & 57.5 \\
        GPT-6 Astra & Visual verify, no retry ($M_{\mathrm{retry}}=0$)
& 69.2 & 73.8 \\

GPT-6 Astra & Visual verify + bounded retry ($M_{\mathrm{retry}}=2$)
& 73.6 & 75.1  \\
        \bottomrule
    \end{tabular*}
\end{table*}
\paragraph{RoboCasa365.}
We evaluate a 50-task subset of RoboCasa365~\cite{robocasa365}, comprising 18 Atomic-Seen, 16 Composite-Seen, and 16 Composite-Unseen tasks, with 50 episodes per task ($N=2{,}500$).
The first five episodes per task ($N=250$) are used for task-level router calibration, while the remaining 45 episodes ($N=2{,}250$) form the held-out evaluation set.
The reported full-benchmark results include all 2,500 episodes, including the 250 router-calibration episodes.
We additionally distinguish the 2,250 episodes held out from router selection for evaluation.
We fine-tune $\pi_{0.5}$ from the official 75k-step checkpoint for 7,500 cumulative steps using 1,437 trajectories from seen tasks (563 human demonstrations and 874 recovery rollouts), excluding all Composite-Unseen tasks.

GPT-6 Astra decomposes the task into up to three ordered manipulation stages.
The task-level routing decisions are determined from five calibration episodes per task and frozen before evaluation.
For Assisted execution, the MFM identifies stage-specific targets and selects either top-down or front-horizontal approaches.
Grounding uses calibrated left and right fixed RGB cameras, supplemented by an actively positioned wrist camera when needed.
The adapted VLA receives three RGB views and proprioceptive states, with $224\times224$ image inputs and replanning every five action steps.
Dynamic dense geometric captions provide stage-specific target-relative guidance during execution.

\paragraph{ALOHA-Sim-PRO.}
ALOHA-Sim-PRO evaluates bimanual cube transfer using a 14-DoF robot.
We use the official $\pi_0$ checkpoint without additional fine-tuning and a fixed 400-step episode horizon.
The benchmark covers five conditions: Original, Semantic, Object, Position, and Task, with 50 seeds per condition.
Semantic perturbations paraphrase the instruction; Object perturbations change the cube to an unseen color; Position perturbations move it outside the training region; and Task perturbations combine target appearance and position changes with an updated instruction.

\begin{figure}[t]
    \centering
    \includegraphics[width=\linewidth]{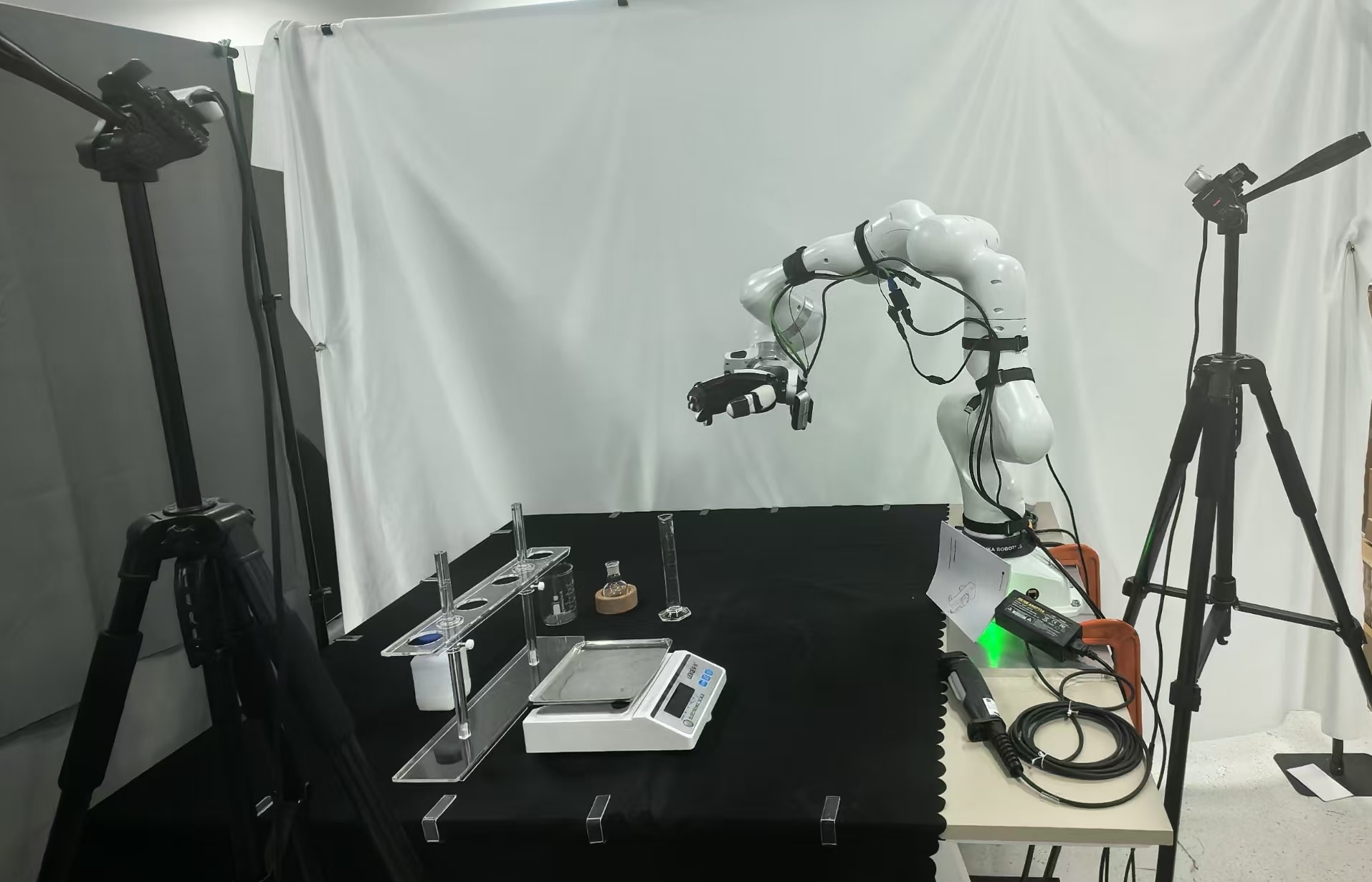}
    \caption{\textbf{Real-world experimental setup.}
    A Franka robotic arm equipped with a dexterous hand operates in a tabletop workspace containing laboratory glassware and an electronic scale. Two external cameras capture the workspace from different viewpoints.}
    \label{fig:real_world_setup}
\end{figure}

\paragraph{Real-World Task.}
\cref{fig:real_world_setup} shows our real-world experimental setup, comprising a Franka robotic arm equipped with a dexterous hand, two external cameras, and a tabletop workspace with laboratory glassware and an electronic scale. 
We evaluate 11 transparent-glassware manipulation tasks using a 7-DoF Franka arm equipped with a 12-DoF XHand dexterous hand.
The perception system consists of one wrist-mounted RGB camera and two tripod-mounted Intel RealSense RGB cameras.
Each task is evaluated with 20 trials per setting.
Scene generalization introduces unseen laboratory distractors, while Position generalization moves targets outside the demonstration region.

We use GPT-6 Astra for task-conditioned grounding and visual grasp verification, together with the adapted VLA for fine-grained manipulation.
Calibrated fixed and wrist RGB views provide two-view geometric measurements of grasp and placement targets.
The resulting 3D targets guide Cartesian free-space motion, while the VLA retains control over local contact-rich operations.
Visual verification and bounded re-grasp recovery are applied when necessary.

\begin{table}[t]
    \centering
    \small
    \setlength{\tabcolsep}{2.2pt}
    \caption{\textbf{GroundingDINO vs.\ MFM backbones}
    on LIBERO-PRO (\texttt{init 5--9}).
    err: median horizontal (x,y) grounding error (cm);
    $\leq 6$\,cm: hit rate (\%).
    Assisted and Overall success (\%) use $250$ and $800$
    episodes, respectively, with the same $417/550$
    successful Direct episodes.}
    \label{tab:supp_gdino}
    \resizebox{\linewidth}{!}{%
    \begin{tabular}{@{}lcccc@{}}
        \toprule
        Backbone & err$\downarrow$ & $\leq6$\,cm$\uparrow$
        & Assist. & Overall \\
        \midrule
        GroundingDINO (top-edge, no verify)
        & 11.97 & 20.9 & 10.0 & 55.3 \\
        GroundingDINO (box center, no verify)
        & 9.04 & 49.1 & 16.4 & 57.3 \\
        GroundingDINO + MFM verify
        & 9.04 & 49.1 & 17.2 & 57.5 \\
        Kimi K3
        & 22.94 & 8.7 & 29.6 & 61.4 \\
        Qwen3-VL-30B
        & 2.25 & 63.0 & 38.0 & 64.0 \\
        Gemini 4 Argon
        & 2.75 & 76.1 & 58.4 & 70.4 \\
        Grok 4.7
        & 0.92 & 91.3 & 59.2 & 70.6 \\
        GPT-5.6
        & 0.57 & 93.5 & 60.4 & 71.0 \\
        GPT-6 Astra
        & \textbf{0.16} & \textbf{97.3}
        & \textbf{73.6} & \textbf{75.1} \\
        \bottomrule
    \end{tabular}%
    }
\end{table}

\section{Additional Experiments}
\label{sec:supp_exp}

\subsection{Replacing the MFM with GroundingDINO}
\label{sec:supp_gdino}

We replace MFM-based grounding with GroundingDINO-Base~\cite{liu2024grounding}, keeping the adapted VLA and suite router fixed.
As shown in~\cref{tab:supp_gdino}, selecting the bounding-box center improves Assisted success over the top-edge midpoint ($16.4\%$ vs.\ $10.0\%$), but remains substantially below GPT-6 Astra ($73.6\%$).
Adding live MFM grasp verification or a simulation target-ID oracle yields only $17.2\%$ success (\cref{tab:supp_gdino_2x2}), suggesting that verification alone cannot compensate for inaccurate grounding.

\subsection{Router Calibration Budget and Margin}
\label{sec:supp_router}

We vary the calibration budget $K$ over all $\binom{5}{K}$ subsets of \texttt{init 0--4} and evaluate on \texttt{init 5--9}.
As shown in~\cref{tab:supp_router_K}, suite-level routing remains stable even with $K{=}1$ ($75.1{\pm}0.2\%$), whereas task-level routing benefits more from additional calibration data, reaching $78.5\%$ at $K{=}5$.
Performance is unchanged across the tested margins from $0$ to $0.10$, while larger margins reduce both assistance frequency and success (\cref{tab:supp_router_delta}).
We use suite-level routing with $K{=}5$ and $\delta{=}0.05$ by default.
The paired episode oracle, which selects between execution modes using evaluation outcomes, achieves $80.9\%$.

\begin{table}[t]
    \centering
    \small
    \setlength{\tabcolsep}{0.5pt}
    \caption{\textbf{Router calibration budget}
    on LIBERO-PRO ($N{=}800$).
    Results cover all $\binom{5}{K}$ calibration subsets.
    Suite/Task denote routing granularity;
    Assist.\ is the Assisted episode rate. All rates are percentages.}
    \label{tab:supp_router_K}
    \begin{tabular*}{\linewidth}
        {@{\extracolsep{\fill}}lcccc@{}}
        \toprule
        $K$ & Suite Succ.(\%) & Assist.(\%) & Task Succ.(\%) & Assist.(\%) \\
        \midrule
        All Direct & 62.6 & 0.0 & 62.6 & 0.0 \\
        All Assisted & 62.3 & 100.0 & 62.3 & 100.0 \\
        \midrule
        1 & $75.1{\pm}0.2$ & 28.8 & $74.6{\pm}0.9$ & 18.2 \\
        2 & $75.2{\pm}0.2$ & 28.1 & $76.1{\pm}0.9$ & 22.6 \\
        3 & $75.2{\pm}0.1$ & 28.8 & $77.1{\pm}0.7$ & 25.0 \\
        4 & $75.2{\pm}0.1$ & 30.0 & $78.0{\pm}0.4$ & 26.9 \\
        5 & 75.1 & 31.2 & \textbf{78.5} & 27.5 \\
        \midrule
        Episode oracle & \multicolumn{4}{c}{80.9} \\
        \bottomrule
    \end{tabular*}
\end{table}

\begin{table}[t]
    \centering
    \small
    \setlength{\tabcolsep}{3pt}
    \caption{\textbf{Routing margin sensitivity}
    at $K{=}5$ ($N{=}800$).
    Assist.\ is the Assisted episode rate.
    All rates are percentages.}
    \label{tab:supp_router_delta}
    \begin{tabular*}{\linewidth}
        {@{\extracolsep{\fill}}lcccc@{}}
        \toprule
        $\delta$ & Suite Succ. (\%) & Assist.(\%)  & Task Succ. (\%) & Assist. (\%) \\
        \midrule
        0.00 & 75.1 & 31.2 & 78.5 & 27.5 \\
        0.05 & 75.1 & 31.2 & 78.5 & 27.5 \\
        0.10 & 75.1 & 31.2 & 78.5 & 27.5 \\
        0.20 & 73.0 & 18.8 & 77.3 & 21.2 \\
        0.40 & 71.9 & 12.5 & 74.8 & 15.6 \\
        \bottomrule
    \end{tabular*}
\end{table}

\begin{table}[t]
    \centering
    \footnotesize
    \setlength{\tabcolsep}{2pt}
    \caption{\textbf{Three action-sampling seeds}
    on LIBERO-PRO (\texttt{init 5--9}, $N{=}800$ per seed).
    Results are reported as
mean ± standard deviation (\%).}
    \label{tab:supp_seeds}
    \begin{tabular}{@{}lccccc@{}}
        \toprule
        Method & Sem. & Obj. & Pos. & Task & Overall \\
        \midrule
        Base $\pi_{0.5}$
        & $93.7{\pm}1.3$
        & $82.3{\pm}0.9$
        & $25.5{\pm}3.5$
        & $25.5{\pm}1.8$
        & $56.8{\pm}0.9$ \\
        MFM-VLA
        & $93.5{\pm}2.2$
        & $82.8{\pm}1.7$
        & $\mathbf{53.3{\pm}2.5}$
        & $\mathbf{59.5{\pm}3.1}$
        & $\mathbf{72.3{\pm}2.0}$ \\
        \bottomrule
    \end{tabular}
\end{table}

\subsection{Multi-Seed Sampling Variance}
\label{sec:supp_seeds}

We evaluate three independent action-sampling seeds on \texttt{init 5--9}, with $800$ episodes per seed and fixed model checkpoints.
As shown in~\cref{tab:supp_seeds}, MFM-VLA improves mean overall success from $56.8\%$ to $72.3\%$.
The gains concentrate on Position and Task perturbations ($+27.8$ and $+34.0$ percentage points), while Semantic and Object performance remains similar, supporting the main findings across sampling seeds.

\subsection{Additional Qualitative Results}
\label{sec:supp_qualitative}

\begin{figure*}[p]
    \centering
    \includegraphics[
        width=\textwidth,
        height=0.84\textheight,
        keepaspectratio
    ]{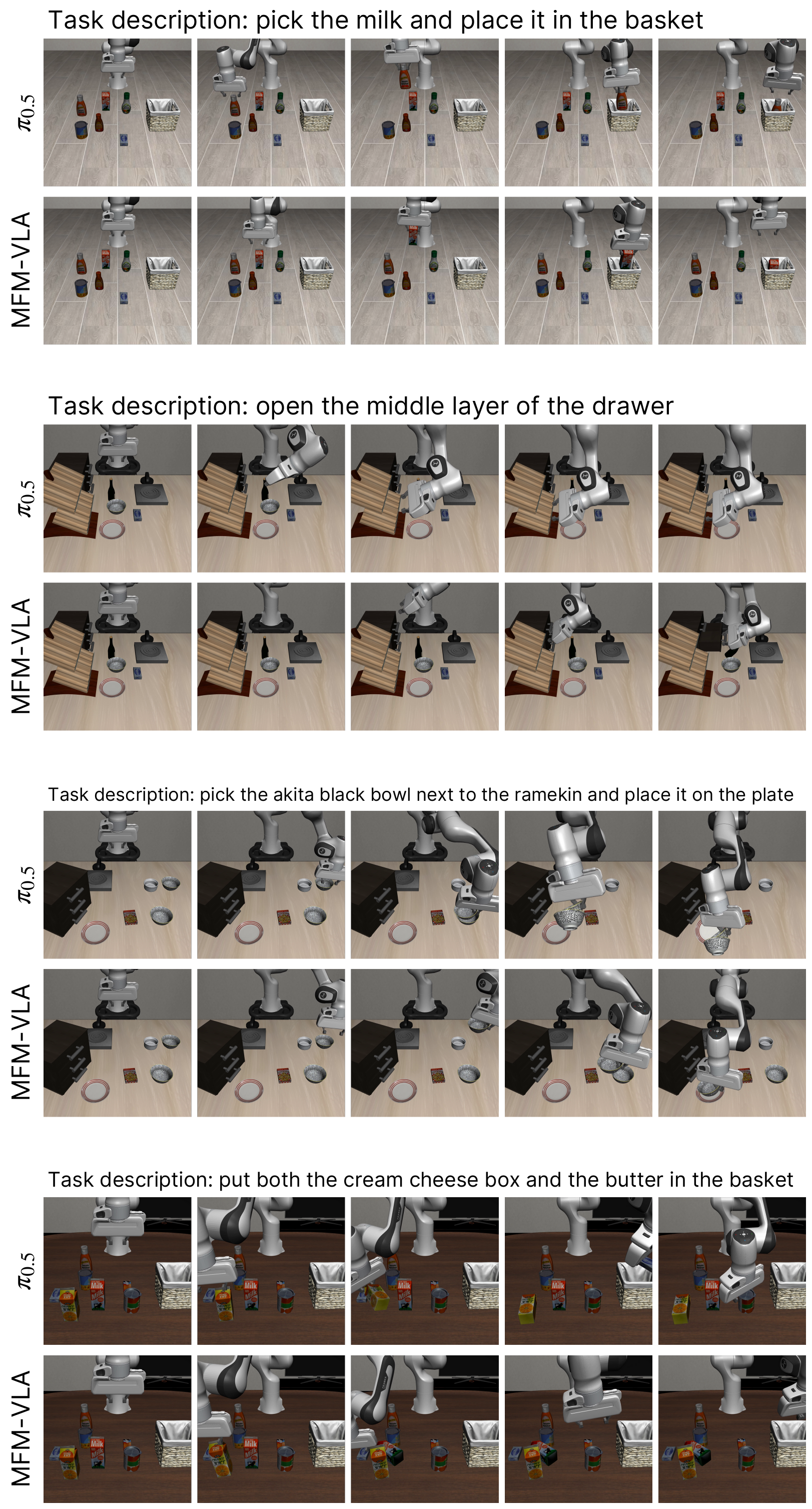}
    \caption{\textbf{Qualitative comparisons on LIBERO-PRO.}
    Examples cover object selection, drawer manipulation, and multi-object rearrangement.
    For each task, the upper and lower rows show $\pi_{0.5}$ and MFM-VLA, respectively. 
    Time progresses from left to right.}
    \label{fig:supp_vis_libero}
\end{figure*}

\begin{figure*}[p]
    \centering
    \includegraphics[
        width=\textwidth,
        height=0.84\textheight,
        keepaspectratio
    ]{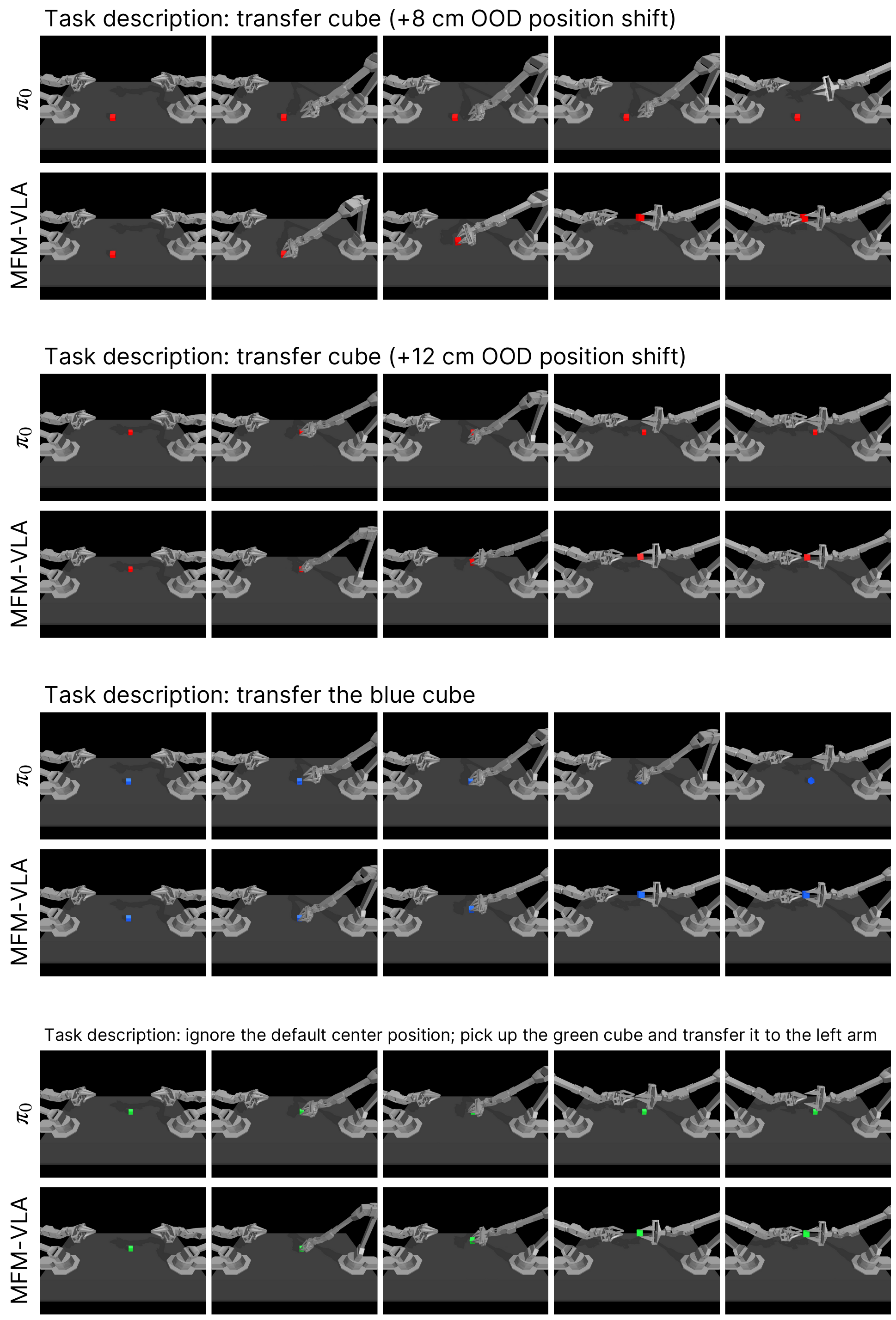}
    \caption{\textbf{Qualitative comparisons on ALOHA-Sim-PRO.}
    Examples show bimanual cube transfer under position shifts and different target instructions.
    For each task, the upper and lower rows show $\pi_0$ and MFM-VLA, respectively. 
    Time progresses from left to right.}
    \label{fig:supp_vis_aloha}
\end{figure*}

\begin{figure*}[p]
    \centering
    \includegraphics[
        width=\textwidth,
        height=0.84\textheight,
        keepaspectratio
    ]{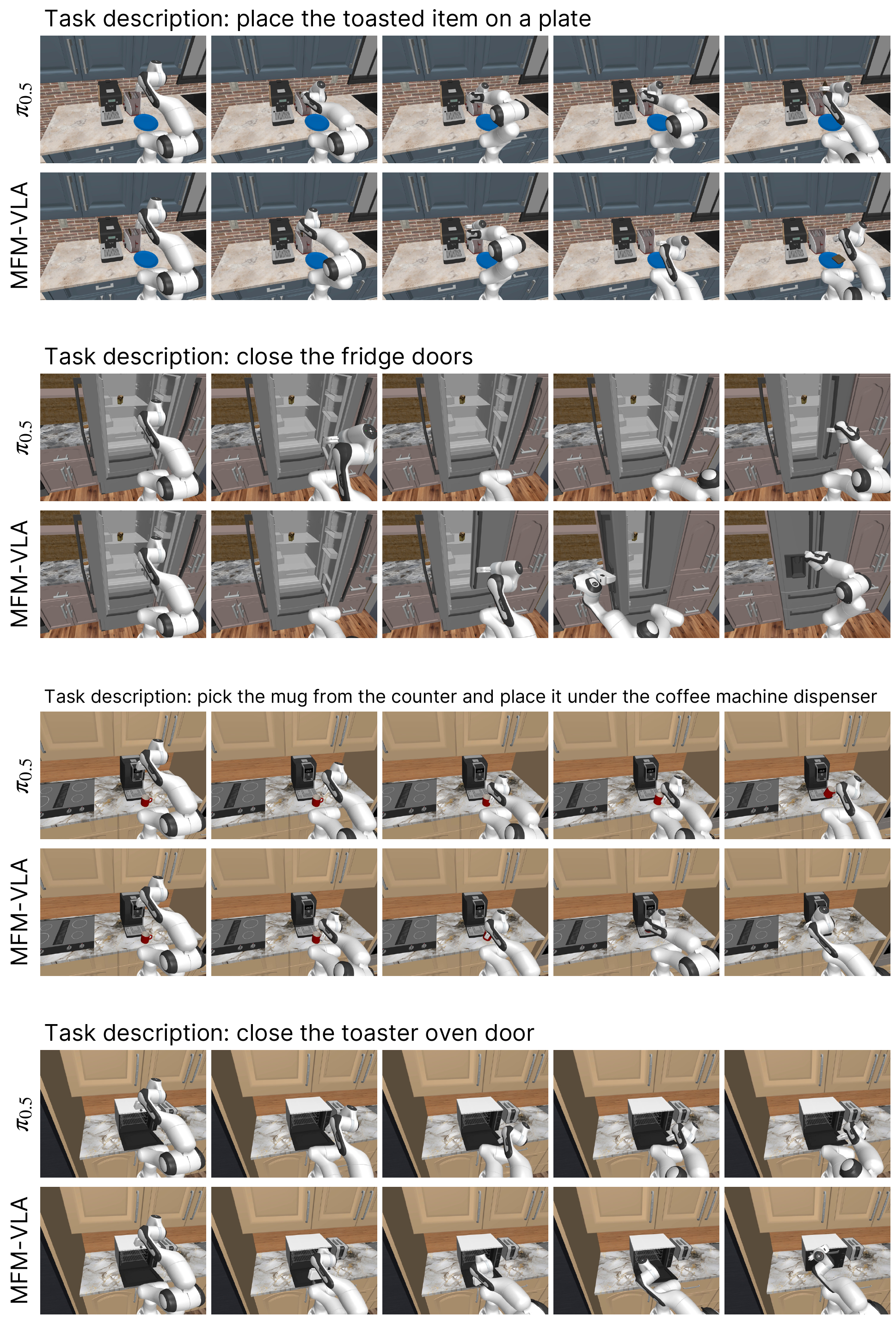}
    \caption{\textbf{Qualitative comparisons on RoboCasa365.}
    Examples cover object placement and toaster oven door manipulation.
    For each task, the upper and lower rows show $\pi_{0.5}$ and MFM-VLA, respectively. 
    Time progresses from left to right.}
    \label{fig:supp_vis_robocasa}
\end{figure*}

\begin{figure*}[p]
    \centering
    \includegraphics[
        width=\textwidth,
        height=0.84\textheight,
        keepaspectratio
    ]{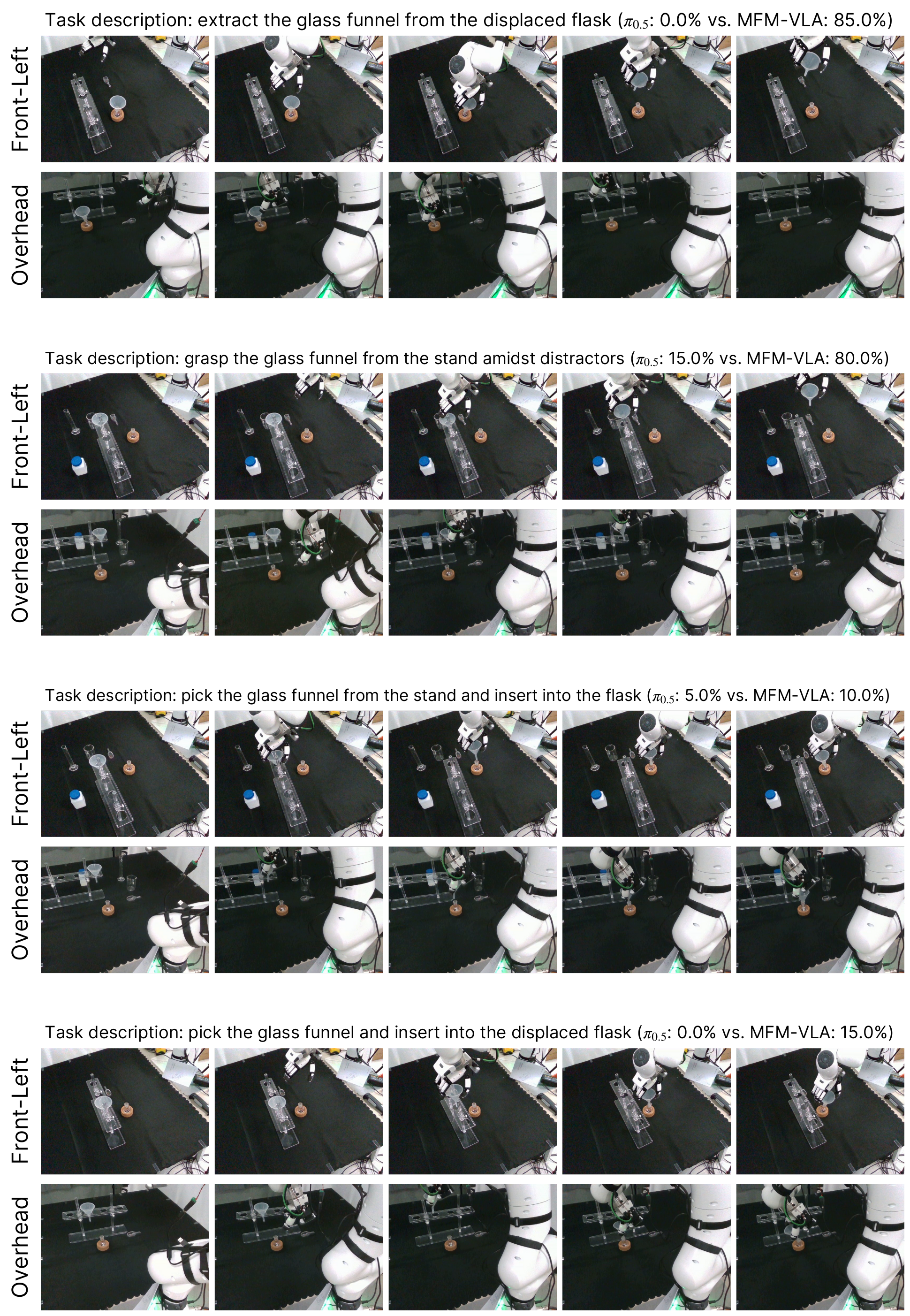}
    \caption{\textbf{Qualitative results on real-world laboratory tasks.}
    Examples show glass funnel extraction, grasping, and insertion with object displacement or distractors.
    For each task, the upper and lower rows show front-left and overhead views, respectively.
    Time progresses from left to right.}
    \label{fig:supp_vis_realworld}
\end{figure*}

We provide additional qualitative results on LIBERO-PRO, ALOHA-Sim-PRO, RoboCasa365, and real-world laboratory tasks in~\cref{fig:supp_vis_libero,fig:supp_vis_aloha,fig:supp_vis_robocasa,fig:supp_vis_realworld}, respectively.
These examples illustrate manipulation behaviors across different target instructions, spatial configurations, and tasks.

\end{document}